\documentclass[10pt, a4paper]{article}
\pdfoutput=1
\usepackage{lrec}
\usepackage{multibib}\newcites{languageresource}{Language Resources}
\usepackage{graphicx}
\usepackage{tabularx}
\usepackage{soul}

\usepackage{titlesec}
\titleformat{\section}{\normalfont\large\bf\center}{\thesection.}{1em}{}
\titleformat{\subsection}{\normalfont\SmallTitleFont\bf\raggedright}{\thesubsection.}{1em}{}
\titleformat{\subsubsection}{\normalfont\normalsize\bf\raggedright}{\thesubsubsection.}{1em}{}
\renewcommand\thesection{\arabic{section}}
\renewcommand\thesubsection{\thesection.\arabic{subsection}}
\renewcommand\thesubsubsection{\thesubsection.\arabic{subsubsection}}

\usepackage[utf8]{inputenc}
\usepackage{hyperref}
\usepackage{xstring}

\newcommand{\citep}[1]{\cite{#1}}
\newcommand{\citet}[1]{\newcite{#1}}
\newenvironment{citemize}{\begin{list}{$\bullet$}{\topsep=\smallskipamount\itemsep=0pt\parsep=1pt\labelwidth=.5em}}{\end{list}}

\usepackage{multirow}
\usepackage{textcomp}

\usepackage{xcolor}
\usepackage{xspace}
\usepackage{pifont}
\newcommand{\YES}{\ding{51}\xspace}
\newcommand{\NO}{{\color{gray!50}\ding{55}}\xspace}

\title{UDPipe at EvaLatin 2020: Contextualized Embeddings\\and Treebank Embeddings}

\name{Milan Straka, Jana Straková}

\address{
  Charles University \\
  Faculty of Mathematics and Physics \\
  Institute of Formal and Applied Linguistics \\
  \{straka,strakova\}@ufal.mff.cuni.cz\\}

\abstract{
  We present our contribution to the EvaLatin shared task, which is the first
  evaluation campaign devoted to the evaluation of NLP tools for Latin.
  We submitted a system based on UDPipe 2.0, one of the winners of the
  CoNLL 2018 Shared Task, The 2018 Shared Task on Extrinsic Parser Evaluation
  and SIGMORPHON 2019 Shared Task. Our system places first by a wide margin both in lemmatization and POS tagging in
  the open modality, where additional supervised data is allowed, in which case
  we utilize all Universal Dependency Latin treebanks. In the closed modality,
  where only the EvaLatin training data is allowed, our system achieves the
  best performance in lemmatization and in classical subtask of POS tagging,
  while reaching second place in cross-genre and cross-time settings.
  In the ablation experiments, we also evaluate the influence of BERT and
  XLM-RoBERTa contextualized embeddings, and the treebank encodings of the
  different flavors of Latin treebanks.
  \\\newline 
}

\begin{document}

\maketitleabstract

\section{Introduction}
\label{section:introduction}

This paper describes our participant system to the EvaLatin 2020 shared
task~\cite{evalatin2020}.
Given a segmented and tokenized text in CoNLL-U format with surface forms as in

\begin{small}
\begin{verbatim}
# sent_id = 1
1  Dum        _          _       _   ...
2  haec       _          _       _   ...
3  in         _          _       _   ...
4  Hispania   _          _       _   ...
5  geruntur   _          _       _   ...
6  C.         _          _       _   ...
7  Trebonius  _          _       _   ...
\end{verbatim}
\end{small}

\noindent the task is to infer lemmas and POS tags:

\begin{small}
\begin{verbatim}
# sent-id = 1
1  Dum        dum        SCONJ   _   ...
2  haec       hic        DET     _   ...
3  in         in         ADP     _   ...
4  Hispania   Hispania   PROPN   _   ...
5  geruntur   gero       VERB    _   ...
6  C.         Gaius      PROPN   _   ...
7  Trebonius  Trebonius  PROPN   _   ...
\end{verbatim}
\end{small}

The EvaLatin 2020 training data consists of 260k words of annotated texts from
five authors. In the closed modality, only the given training data may be used,
while in open modality any additional resources can be utilized.

We submitted a system based on UDPipe 2.0~\citep{UDPipe-SIGMORPHON2019}.
In the open modality, our system also uses all three
UD~2.5~\citelanguageresource{UD2.5} Latin treebanks as additional training data
and places first by a wide margin both in lemmatization and POS tagging.

In the closed modality, our system achieves the best performance in
lemmatization and in classical subtask of POS tagging (consisting of texts of
the same five authors as the training data), while reaching
second place in cross-genre and cross-time setting.

Additionally, we evaluated the effect of:
\begin{citemize}
  \item BERT~\citep{BERT} and XLM-RoBERTa~\citep{XLM-R} contextualized embeddings;
  \item various granularity levels of treebank embeddings~\citep{stymne-etal-2018-parser}.
\end{citemize}

\section{Related Work}
\label{section:related_work}

The EvaLatin 2020 shared task~\cite{evalatin2020} is reminiscent of the SIGMORPHON2019
Shared Task~\cite{SIGMORPHON2019}, where the goal was also to perform
lemmatization and POS tagging, but on 107 corpora in 66 languages.
It is also related to CoNLL 2017 and 2018 Multilingual Parsing from Raw Texts
to Universal Dependencies shared
tasks~\citep{UDST2017:overview,UDST2018:overview}, in which the goal was to
process raw texts into tokenized sentences with POS tags, lemmas, morphological
features and dependency trees of the Universal Dependencies project~\citep{ud},
which seeks to develop cross-linguistically consistent treebank
annotation of morphology and syntax for many languages.

UDPipe~2.0~\citep{UDPipe2016,UDPipe2018} was one of the
winning systems of the CoNLL 2018 shared task, performing the POS tagging,
lemmatization and dependency parsing jointly. Its
modification~\citep{UDPipe-SIGMORPHON2019} took part in the SIGMORPHON 2019
shared task, delivering best performance in lemmatization and comparable
to best performance in POS tagging.

A new type of deep contextualized word representation was introduced by
\citet{Peters2018}. The proposed embeddings, called ELMo, were obtained from
internal states of deep bidirectional language model, pretrained on a large
text corpus. The idea of ELMos was extended to BERT by \citet{BERT}, who instead of
a bidirectional recurrent language model employ a Transformer
\hbox{\citep{vaswani:2017}} architecture. A multilingual BERT model trained on 102
languages can significantly improve performance in many NLP tasks across many
languages. Recently, XLM-RoBERTa, an improved multilingual model based on BERT,
was proposed by~\citet{XLM-R}, which appears to offer stronger performance in
multilingual representation~\citep{XLM-R,MLQA}.

\section{Methods}
\label{section:methods}

\subsection{Architecture Overview}

\begin{figure}[t]
  \begin{center}
    \includegraphics[width=.9\hsize]{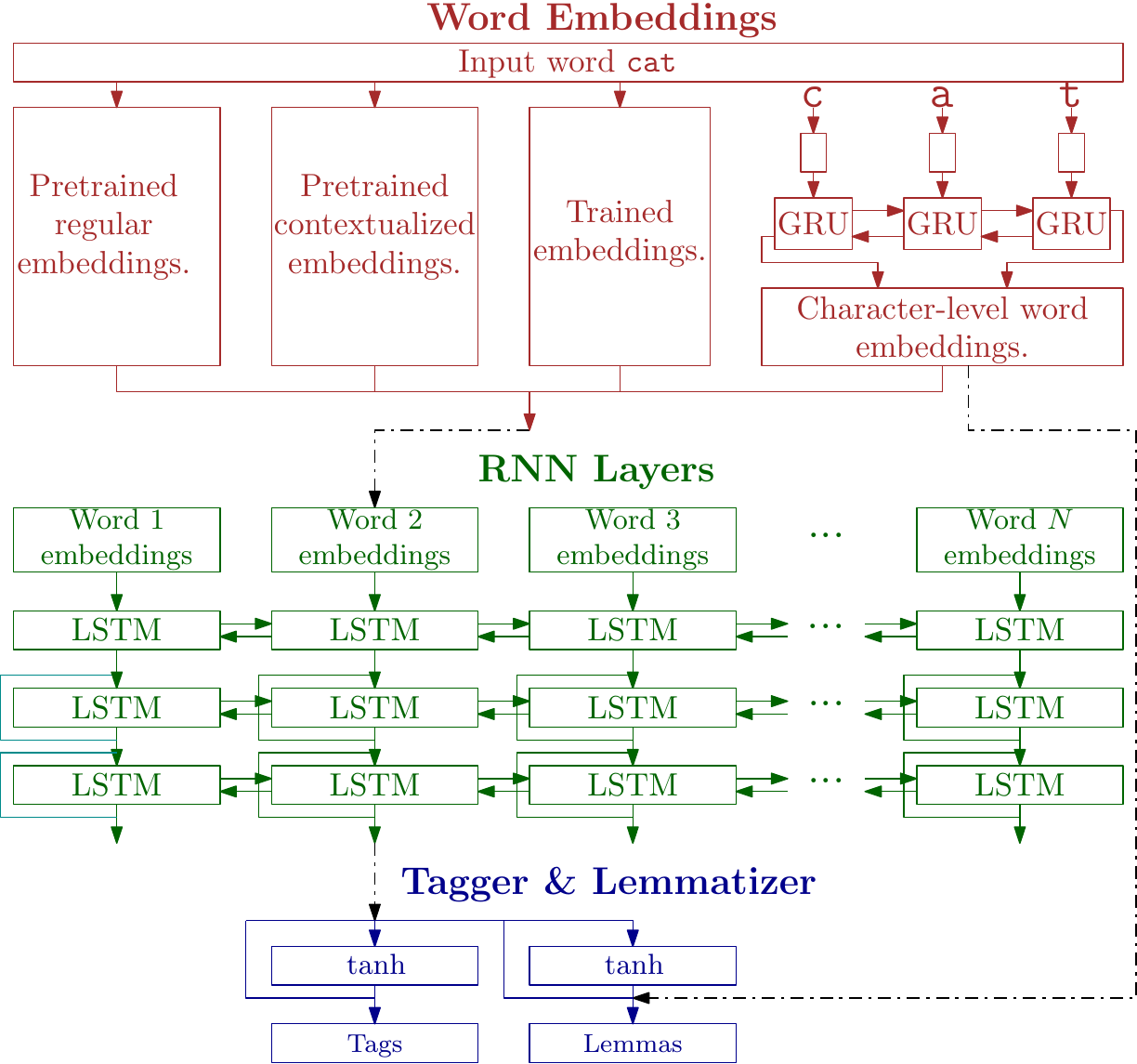}
    \caption{The UDPipe network architecture of the joint tagger and lemmatizer.}
    \label{fig:architecture}
  \end{center}
\end{figure}

Our architecture is based on UDPipe entry to SIGMORPHON
2019 Shared Task~\cite{UDPipe-SIGMORPHON2019}, which is available at
\url{https://github.com/ufal/sigmorphon2019}. The resulting model is presented
in Figure~\ref{fig:architecture}.

In short, the architecture is a multi-task model predicting jointly lemmas
and POS tags. After embedding input words, three shared bidirectional
LSTM~\citep{Hochreiter:1997:LSTM} layers are performed. Then, softmax
classifiers process the output and generate the lemmas and POS tags.

The lemmas are generated by classifying into a set of edit scripts which
process input word form and produce lemmas by performing character-level edits
on the word prefix and suffix. The lemma classifier additionally takes the
character-level word embeddings as input. The lemmatization is further
described in Section~\ref{section:lemmatization}.

The input word embeddings are the same as in the previous versions
of UDPipe~2.0:

\begin{citemize}
  \item \textbf{end-to-end word embeddings},
  \item \textbf{character-level word embeddings:} We employ bidirectional
    GRUs \cite{Cho2014,Graves2005} of dimension $256$ in line with
    \cite{Ling2015}: we represent every Unicode character with a vector of
    dimension $256$, and concatenate GRU output for forward and reversed word
    characters. The character-level word embeddings are trained together with
    UDPipe network.
  \item \textbf{pretrained word embeddings:} We use FastText word embeddings
    \cite{FastText} of dimension $300$, which we pretrain
    on plain texts provided by CoNLL 2017 UD Shared
    Task~\citelanguageresource{CoNLL2017RawTexts}, using segmentation and
    tokenization trained from the UD data.\footnote{We use
    {\scriptsize\ttfamily-minCount 5 -epoch 10 -neg 10} options.}
  \item \textbf{pretrained contextualized word embeddings:} We use the
    {\small\ttfamily Multilingual Base Uncased} 
    BERT~\cite{BERT} model to provide contextualized embeddings of
    dimensionality $768$, averaging the last layer of subwords belonging to
    the same word.
\end{citemize}

We refer the readers for detailed description of the architecture and the
training procedure to \citet{UDPipe-SIGMORPHON2019}.

\subsection{Lemmatization}
\label{section:lemmatization}

\begin{table*}[t]
  \begin{center}
    \small
    \tabcolsep=2pt
    \begin{tabular}{l||l|l||l}
      \multicolumn{1}{c||}{Lemma Rule} & \multicolumn{1}{c|}{Casing Script}
        & \multicolumn{1}{c||}{Edit Script} & \multicolumn{1}{c}{Most Frequent Examples} \\\hline\hline
        \verb|↓0;d¦|        & all lowercase           & do nothing & et→et, in→in, non→non, ut→ut, ad→ad \\\hline
        \verb|↓0;d¦-+u+s|   & all lowercase           & change last char to \verb|us| & suo→suus, loco→locus, Romani→romanus, sua→suus \\\hline
        \verb|↓0;d¦---+o|   & all lowercase           & change last 3 chars to \verb|o| & dare→do, dicere→dico, fieri→fio, uidetur→uideo, data→do \\\hline
        \verb|↓0;d¦-+s|     & all lowercase           & change last char to \verb|s| & quid→quis, id→is, rei→res, omnia→omnis, rem→res \\\hline
        \verb|↓0;d¦----+o|  & all lowercase           & change last 4 chars to \verb|o| & hominum→homo, dedit→do, homines→homo \\\hline
        \verb|↓0;d¦--+o|    & all lowercase           & change last 2 chars to \verb|o| & habere→habeo, dicam→dico, ferre→fero, dat→do \\\hline
        \verb|↓0;d¦--+u+s|  & all lowercase           & change last 2 chars to \verb|us| & publicae→publicus, suis→suus, suam→suus, suos→suus \\\hline
        \verb|↓0;d¦-|       & all lowercase           & remove last character & gratiam→gratia, causam→causa, uitam→uita, copias→copia \\\hline
        \verb|↓0;d¦-+u+m|   & all lowercase           & change last char to \verb|um| & belli→bellum, posse→possum, bello→bellum \\\hline
        \verb|↓0;d¦---+s|   & all lowercase           & change last 3 chars to \verb|s| & omnibus→omnis, rebus→res, nobis→nos, rerum→res \\\hline
        \verb|↑0¦↓1;d¦|     & $1^\mathrm{st}$ upper, then lower & do nothing & Caesar→Caesar, Plinius→Plinius, Antonius→Antonius \\\hline
        \verb|↓0;d¦-----+o| & all lowercase           & change last 5 chars to \verb|o| & uideretur→uideo, uidebatur→uideo, faciendum→facio \\\hline
        \verb|↓0;d¦--+i|    & all lowercase           & change last 2 chars to \verb|i| & quod→qui, quae→qui, quem→qui, quos→qui, quam→qui \\\hline
        \verb|↓0;d¦---|     & all lowercase           & remove last 3 characters & quibus→qui, legiones→legio, legionum→legio, legionis→legio \\\hline
        \verb|↓0;d¦--+s|    & all lowercase           & change last 2 chars to \verb|s| & omnium→omnis, hostium→hostis, parte→pars, urbem→urbs \\\hline
        \dots               & \dots                   & \dots & \dots \\\hline
        \verb|↓0;ais|       & all lowercase           & ignore form, use \verb|is| & eum→is, eo→is, ea→is, eorum→is, eam→is \\\hline
    \end{tabular}
  \end{center}
  \caption{Fifteen most frequent lemma rules in EvaLatin training data ordered
  from the most frequent one, and the most frequent rule with an absolute edit
  script.}
  \label{table:lemmarules}
\end{table*}

The lemmatization is modeled as a multi-class classification, in which the
classes are the complete rules which lead from input forms to the lemmas. We
call each class encoding a transition from input form to lemma a \textit{lemma
rule}. We create a lemma rule by firstly encoding the correct casing as
a \textit{casing script} and secondly by creating a sequence of character
edits, an \textit{edit script}.

Firstly, we deal with the casing by creating a \textit{casing script}. By
default, word form and lemma characters are treated as lowercased. If the lemma
however contains upper-cased characters, a rule is added to the casing script
to uppercase the corresponding characters in the resulting lemma. For example,
the most frequent casing script is ``keep the lemma lowercased (don't do
anything)'' and the second most frequent casing script is ``uppercase the first
character and keep the rest lowercased''.

As a second step, an \textit{edit script} is created to convert input
lowercased form to lowercased lemma. To ensure meaningful editing, the form is split to three parts,
which are then processed separately: a prefix, a root (stem) and a suffix. The
root is discovered by matching the longest substring shared between the form
and the lemma; if no matching substring is found (e.g., form \textit{eum} and
lemma \textit{is}), we consider the word irregular, do not process it with any
edits and directly replace the word form with the lemma. Otherwise, we proceed
with the edit scripts, which process the prefix and the suffix separately and
keep the root unchanged. The allowed character-wise operations are character
copy, addition and deletion.

The resulting \textit{lemma rule} is a concatenation of a casing script and an
edit script. The most common lemma rules present in EvaLatin training data are presented
in Table~\ref{table:lemmarules}.

Using the generated lemma rules, the task of lemmatization is then reduced to
a multiclass classification task, in which the artificial neural network
predicts the correct lemma rule.

\subsection{Treebank Embedding}
\label{section:treebank_embedding}

In the open modality, we additionally train on all three UD 2.5 Latin treebanks. In
order to recognize and handle possible differences in the treebank annotations,
we employ treebank embeddings following~\cite{stymne-etal-2018-parser}.

Furthermore, given that the author name is a known information both during
training and prediction time, we train a second model with
author-specific embeddings for the individual authors. We employ the model with
author-specific embeddings whenever the predicted text comes from one of the
training data authors (in-domain setting) and a generic model otherwise
(out-of-domain setting).

\section{Results}
\label{section:results}

\begin{table}[t]
  \begin{center}
    \small
    \tabcolsep=4.5pt
    \catcode`@ = 13\def@{\bfseries}
    \catcode`! = 13\def!{\itshape}
    \catcode`? = 13\def?{\hphantom{(0)}}
    \begin{tabular}{l||r|r|r}
      \multirow{2}{*}{System} & \multicolumn{3}{c}{Lemmatization} \\\cline{2-4}
      & classical & cross-genre & cross-time \\\hline
     @UDPipe -- open   & 96.19 (1) & 87.13 (1) & 91.01 (1) \\\hline
     @UDPipe -- closed & 95.90 (2) & 85.47 (3) & 87.69 (2) \\\hline
      P2 -- closed 1   & 94.76 (3) & 85.49 (2) & 85.75 (3) \\\hline
      P3 -- closed 1   & 94.60 (4) & 81.69 (5) & 83.92 (4) \\\hline
      P2 -- closed 2   & 94.22 (5) & 82.69 (4) & 83.76 (5) \\\hline
      \hline
     !Post ST -- open  &!96.35  ?  &!87.48  ?  &!91.07  ?  \\\hline
     !Post ST -- closed&!95.93  ?  &!85.94  ?  &!87.88  ?  \\\hline
    \end{tabular}
  \end{center}
  \caption{Official ranking of EvaLatin lemmatization.
  Additionally, we include our best post-competition model in italic.}
  \label{table:official_lemmatization}
\end{table}

\begin{table}[t]
  \begin{center}
    \small
    \tabcolsep=4.5pt
    \catcode`@ = 13\def@{\bfseries}
    \catcode`! = 13\def!{\itshape}
    \catcode`? = 13\def?{\hphantom{(0)}}
    \begin{tabular}{l||r|r|r}
      \multirow{2}{*}{System} & \multicolumn{3}{c}{Tagging} \\\cline{2-4}
      & classical & cross-genre & cross-time \\\hline
     @UDPipe -- open   & 96.74 (1) & 91.11 (1) & 87.69 (1) \\\hline
     @UDPipe -- closed & 96.65 (2) & 90.15 (3) & 84.93 (3) \\\hline
      P4 -- closed 2   & 96.34 (3) & 90.64 (2) & 87.00 (2) \\\hline
      P3 -- closed 1   & 95.52 (4) & 88.54 (4) & 83.96 (4) \\\hline
      P4 -- closed 3   & 95.35 (5) & 86.95 (6) & 81.38 (7) \\\hline
      P2 -- closed 1   & 94.15 (6) & 88.40 (5) & 82.62 (6) \\\hline
      P4 -- closed 1   & 93.24 (7) & 83.88 (7) & 82.99 (5) \\\hline
      P2 -- closed 2   & 92.98 (8) & 82.93 (8) & 80.78 (8) \\\hline
      P5 -- closed 1   & 90.65 (9) & 73.47 (9) & 76.62 (9) \\\hline
      \hline
     !Post ST -- open  &!96.82  ?  &!91.46  ?  &!87.91  ?  \\\hline
     !Post ST -- closed&!96.76  ?  &!90.50  ?  &!84.70  ?  \\\hline
    \end{tabular}
  \end{center}
  \caption{Official ranking of EvaLatin lemmatization. Additionally, we include
  our best post-competition model in italic.}
  \label{table:official_tagging}
\end{table}

The official overall results are presented in
Table~\ref{table:official_lemmatization} for lemmatization
and in Table~\ref{table:official_tagging} for POS tagging.
In the open modality, our system places first by a wide margin both in
lemmatization and POS tagging. In the closed modality, our system achieves
best performance in lemmatization and in classical subtask of POS tagging
(where the texts from the training data authors are annotated),
and second place in cross-genre and cross-time settings.

\section{Ablation Experiments}
\label{section:ablations}

\begin{table*}[t]
  \begin{center}
    \tabcolsep=4pt
    \small
    \begin{tabular}{l|l|l||r|r|r||r|r|r}
      \multicolumn{1}{c|}{Word} & \multicolumn{1}{c|}{BERT} & \multicolumn{1}{c|}{XLM-RoBERTa} & \multicolumn{3}{c||}{Lemmatization} & \multicolumn{3}{c}{Tagging} \\\cline{4-9}
      \multicolumn{1}{c|}{embeddings} & \multicolumn{1}{c|}{embeddings} & \multicolumn{1}{c|}{embeddings} & classical & cross-genre & cross-time & classical & cross-genre & cross-time \\\hline
      \hline \multicolumn{9}{c}{Open modality} \\\hline
      \NO  & \NO  & \NO  & 96.04 & 86.85 & 90.58   & 96.46 & 90.44 & 87.66 \\\hline
      \YES & \NO  & \NO  & 96.27 & 87.28 & 90.80   & 96.64 & 91.16 & 87.78 \\\hline
      \NO  & \YES & \NO  & 96.19 & 86.76 & 90.78   & 96.70 & 90.34 & 87.50 \\\hline
      \NO  & \NO  & \YES & 96.33 & 86.48 & 90.95   & 96.80 & 90.67 & 87.79 \\\hline
      \YES & \YES & \NO  & 96.28 & 87.28 & 90.80   & 96.74 & 91.11 & 87.69 \\\hline
      \YES & \NO  & \YES & 96.35 & 87.48 & 91.07   & 96.82 & 91.46 & 87.91 \\\hline
      \hline \multicolumn{9}{c}{Closed modality} \\\hline
      \NO  & \NO  & \NO  & 95.62 & 84.62 & 87.63   & 96.14 & 88.90 & 83.59 \\\hline
      \YES & \NO  & \NO  & 95.79 & 85.55 & 88.37   & 96.44 & 90.59 & 84.14 \\\hline
      \NO  & \YES & \NO  & 95.65 & 84.76 & 87.58   & 96.44 & 89.08 & 84.84 \\\hline
      \NO  & \NO  & \YES & 95.93 & 84.97 & 87.63   & 96.67 & 89.36 & 84.24 \\\hline
      \YES & \YES & \NO  & 95.96 & 85.52 & 88.04   & 96.65 & 90.15 & 84.93 \\\hline
      \YES & \NO  & \YES & 95.93 & 85.94 & 87.88   & 96.76 & 90.50 & 84.70 \\\hline
    \end{tabular}
  \end{center}
  \caption{The evaluation of various pretrained embeddings (FastText word
  embeddings, Multilingual BERT embeddings, XLM-RoBERTa embeddings) on the
  lemmatization and POS tagging.}
  \label{table:ablation_embeddings}

  \begin{center}
    \tabcolsep=5pt
    \small
    \begin{tabular}{l||r|r|r||r|r|r}
      & \multicolumn{3}{c||}{Lemmatization} & \multicolumn{3}{c}{Tagging} \\\cline{2-7}
      & classical & cross-genre & cross-time & classical & cross-genre & cross-time \\\hline
      \begin{tabular}{l}The improvement of open modality,\\i.e., using all three UD Latin treebanks\end{tabular}
        & +0.430 & +1.795 & +2.975 & +0.177 & +1.100 & +3.315 \\\hline
    \end{tabular}
  \end{center}
  \caption{The average percentage point improvement in the open modality
  settings compared to the closed modality. The results are averaged over all
  models in Table~\ref{table:ablation_embeddings}.}
  \label{table:ablation_open_closed}
\end{table*}

\begin{table*}[t]
  \begin{center}
    \tabcolsep=4.8pt
    \small
    \begin{tabular}{l||r|r|r||r|r|r}
      & \multicolumn{3}{c||}{Lemmatization} & \multicolumn{3}{c}{Tagging} \\\cline{2-7}
      & \llap{c}lassical & \llap{c}ross-genre & \llap{c}ross-time & \llap{c}lassical & \llap{c}ross-genre & \llap{c}ross-time \\\hline
      Per-author embeddings, per-UD-treebank embeddings & 96.28 & 87.28 & 90.80   & 96.74 & 91.11 & 87.69 \\\hline
      Single EvaLatin embedding, per-UD-treebank embeddings      & 96.28 & 87.28 & 90.80   & 96.70 & 91.11 & 87.69 \\\hline
      Single EvaLatin embedding, single UD-treebank embedding    & 96.23 & 87.22 & 90.78   & 96.68 & 91.14 & 87.63 \\\hline
      EvaLatin and UD treebanks merged           & 96.18 & 87.23 & 90.77   & 96.52 & 91.01 & 86.12 \\\hline
    \end{tabular}
  \end{center}
  \caption{The effect of various kinds of treebank embeddings in open
  modality -- whether the individual authors in EvaLatin get a different or
  the same treebank embedding, and whether the UD treebanks get a different
  treebank embedding, same treebank embedding but different from the EvaLatin
  data, or the same treebank embedding as EvaLatin data.}
  \label{table:ablation_open_variants}
\end{table*}

\begin{table}[t]
  \begin{center}
    \tabcolsep=5pt
    \small
    \begin{tabular}{l|r|r}
      & \multicolumn{1}{c|}{Lemmatization} & \multicolumn{1}{c}{~~~~~Tagging} \\\cline{2-3}
      & \llap{c}lassical & \llap{c}lassical \\\hline
      \begin{tabular}{l}The improvement of\\using per-author\\treebank embeddings\end{tabular} & 0.027 & 0.043 \\\hline
    \end{tabular}
  \end{center}
  \caption{The average percentage point improvement of using per-author treebank embedding
  compared to not distinguishing among authors of EvaLatin data, averaged over
  all models in Table~\ref{table:ablation_embeddings}.}
  \label{table:ablation_per_author}
\end{table}

The effect of various kinds contextualized embeddings is evaluated in
Table~\ref{table:ablation_embeddings}. While BERT embeddings yield only a minor
accuracy increase, which is consistent with~\cite{UDPipeBert} for Latin,
using XLM-RoBERTa leads to larger accuracy improvement.
For comparison, we include the post-competition system with XLM-RoBERTa
embeddings in Tables~\ref{table:official_lemmatization}
and~\ref{table:official_tagging}.

To quantify the boost of the additional training data in the open modality, we
considered all models from the above mentioned
Table~\ref{table:ablation_embeddings}, arriving at the average improvement
presented in Table~\ref{table:ablation_open_closed}. While the performance on
the in-domain test set (classical subtask) improves only slightly, the
out-of-domain test sets (cross-genre and cross-time subtasks) show more
substantial improvement with the additional training data.

The effect of different granularity of treebank embeddings in open modality is
investigated in Table~\ref{table:ablation_open_variants}. When treebank
embeddings are removed from our competition system, the performance deteriorates the most,
even if only a little in absolute terms. This indicates that the UD and
EvaLatin annotations are very consistent. Providing one embedding
for EvaLatin data and another for all UD treebanks improves the performance,
and more so if three UD treebank specific embeddings are used.

Lastly, we evaluate the effect of the per-author embeddings. While on the
development set the improvement was larger, the results on the test sets
are nearly identical. To get more accurate estimate, we computed the average improvement for
all models in Table~\ref{table:ablation_embeddings}, arriving at marginal
improvements in Table~\ref{table:ablation_per_author}, which indicates that
per-author embeddings have nearly no effect on the final system
performance (compared to EvaLatin and UD specific embeddings).

\section{Conclusion}

We described our entry to the EvaLatin 2020 shared task, which placed first
in the open modality and delivered strong performance in the closed modality.

For a future shared task, we think it might be interesting to include also
segmentation and tokenization or extend the shared task with an extrinsic
evaluation.

\section{Acknowledgements}

This work was supported by the grant no. GX20-16819X of the Grant Agency of the
Czech Republic, and has been using language resources stored and distributed by
the LINDAT/CLARIAH-CZ project of the Ministry of Education, Youth and Sports of
the Czech Republic (project LM2018101).

\section{Bibliographical References}
\bibliographystyle{lrec}
\bibliography{evalatin_udpipe}

\begin{thebibliography}{}

\bibitem[\protect\citename{Bojanowski \bgroup et al.\egroup }2017]{FastText}
Bojanowski, P., Grave, E., Joulin, A., and Mikolov, T.
\newblock (2017).
\newblock {Enriching Word Vectors with Subword Information}.
\newblock {\em Transactions of the Association for Computational Linguistics},
  5:135--146.

\bibitem[\protect\citename{Cho \bgroup et al.\egroup }2014]{Cho2014}
Cho, K., van Merrienboer, B., Bahdanau, D., and Bengio, Y.
\newblock (2014).
\newblock On the {P}roperties of {N}eural {M}achine {T}ranslation:
  {E}ncoder-{D}ecoder {A}pproaches.
\newblock {\em CoRR}.

\bibitem[\protect\citename{Conneau \bgroup et al.\egroup }2019]{XLM-R}
Conneau, A., Khandelwal, K., Goyal, N., Chaudhary, V., Wenzek, G.,
  Guzm{\'{a}}n, F., Grave, E., Ott, M., Zettlemoyer, L., and Stoyanov, V.
\newblock (2019).
\newblock {Unsupervised Cross-lingual Representation Learning at Scale}.
\newblock {\em CoRR}, abs/1911.02116.

\bibitem[\protect\citename{Devlin \bgroup et al.\egroup }2019]{BERT}
Devlin, J., Chang, M.-W., Lee, K., and Toutanova, K.
\newblock (2019).
\newblock {BERT}: Pre-training of deep bidirectional transformers for language
  understanding.
\newblock In {\em Proceedings of the 2019 Conference of the North {A}merican
  Chapter of the Association for Computational Linguistics: Human Language
  Technologies, Volume 1 (Long and Short Papers)}, pages 4171--4186,
  Minneapolis, Minnesota, June. Association for Computational Linguistics.

\bibitem[\protect\citename{Graves and Schmidhuber}2005]{Graves2005}
Graves, A. and Schmidhuber, J.
\newblock (2005).
\newblock Framewise phoneme classification with bidirectional lstm and other
  neural network architectures.
\newblock {\em Neural Networks}, pages 5--6.

\bibitem[\protect\citename{Hochreiter and
  Schmidhuber}1997]{Hochreiter:1997:LSTM}
Hochreiter, S. and Schmidhuber, J.
\newblock (1997).
\newblock Long {S}hort-{T}erm {M}emory.
\newblock {\em Neural Comput.}, 9(8):1735--1780, November.

\bibitem[\protect\citename{Lewis \bgroup et al.\egroup }2019]{MLQA}
Lewis, P., Oguz, B., Rinott, R., Riedel, S., and Schwenk, H.
\newblock (2019).
\newblock Mlqa: Evaluating cross-lingual extractive question answering.
\newblock {\em ArXiv}, abs/1910.07475.

\bibitem[\protect\citename{Ling \bgroup et al.\egroup }2015]{Ling2015}
Ling, W., Lu{\'{i}}s, T., Marujo, L., Astudillo, R.~F., Amir, S., Dyer, C.,
  Black, A.~W., and Trancoso, I.
\newblock (2015).
\newblock Finding {F}unction in {F}orm: {C}ompositional {C}haracter {M}odels
  for {O}pen {V}ocabulary {W}ord {R}epresentation.
\newblock {\em CoRR}.

\bibitem[\protect\citename{McCarthy \bgroup et al.\egroup
  }2019]{SIGMORPHON2019}
McCarthy, A.~D., Vylomova, E., Wu, S., Malaviya, C., Wolf-Sonkin, L., Nicolai,
  G., Kirov, C., Silfverberg, M., Mielke, S.~J., Heinz, J., Cotterell, R., and
  Hulden, M.
\newblock (2019).
\newblock {The SIGMORPHON 2019 Shared Task: Morphological Analysis in Context
  and Cross-Lingual Transfer for Inflection}.
\newblock In {\em Proceedings of the 16th Workshop on Computational Research in
  Phonetics, Phonology, and Morphology}, pages 229--244, Florence, Italy,
  August. Association for Computational Linguistics.

\bibitem[\protect\citename{Nivre \bgroup et al.\egroup }2016]{ud}
Nivre, J., de~Marneffe, M.-C., Ginter, F., Goldberg, Y., Haji{\v{c}}, J.,
  Manning, C., McDonald, R., Petrov, S., Pyysalo, S., Silveira, N., Tsarfaty,
  R., and Zeman, D.
\newblock (2016).
\newblock {Universal Dependencies} v1: A multilingual treebank collection.
\newblock In {\em Proceedings of the 10th International Conference on Language
  Resources and Evaluation ({LREC} 2016)}, pages 1659--1666, Portorož,
  Slovenia. European Language Resources Association.

\bibitem[\protect\citename{Peters \bgroup et al.\egroup }2018]{Peters2018}
Peters, M., Neumann, M., Iyyer, M., Gardner, M., Clark, C., Lee, K., and
  Zettlemoyer, L.
\newblock (2018).
\newblock {Deep Contextualized Word Representations}.
\newblock In {\em Proceedings of the 2018 Conference of the North American
  Chapter of the Association for Computational Linguistics: Human Language
  Technologies, Volume 1 (Long Papers)}, pages 2227--2237. Association for
  Computational Linguistics.

\bibitem[\protect\citename{Sprugnoli \bgroup et al.\egroup }2020]{evalatin2020}
Sprugnoli, R., Passarotti, M., Cecchini, F.~M., and Pellegrini, M.
\newblock (2020).
\newblock Overview of the evalatin 2020 evaluation campaign.
\newblock In Rachele Sprugnoli et~al., editors, {\em Proceedings of the LT4HALA
  2020 Workshop - 1st Workshop on Language Technologies for Historical and
  Ancient Languages, satellite event to the Twelfth International Conference on
  Language Resources and Evaluation (LREC 2020)}, Paris, France, May. European
  Language Resources Association (ELRA).

\bibitem[\protect\citename{Straka \bgroup et al.\egroup }2016]{UDPipe2016}
Straka, M., Haji\v{c}, J., and Strakov\'{a}, J.
\newblock (2016).
\newblock {UDPipe: Trainable Pipeline for Processing CoNLL-U Files Performing
  Tokenization, Morphological Analysis, {POS} Tagging and Parsing}.
\newblock In {\em Proceedings of the 10th International Conference on Language
  Resources and Evaluation ({LREC} 2016)}, Portorož, Slovenia. European
  Language Resources Association.

\bibitem[\protect\citename{Straka \bgroup et al.\egroup
  }2019a]{UDPipe-SIGMORPHON2019}
Straka, M., Strakov{\'a}, J., and Hajic, J.
\newblock (2019a).
\newblock {UDPipe at SIGMORPHON 2019: Contextualized Embeddings, Regularization
  with Morphological Categories, Corpora Merging}.
\newblock In {\em Proceedings of the 16th Workshop on Computational Research in
  Phonetics, Phonology, and Morphology}, pages 95--103, Florence, Italy,
  August. Association for Computational Linguistics.

\bibitem[\protect\citename{{Straka} \bgroup et al.\egroup }2019b]{UDPipeBert}
{Straka}, M., {Strakov{\'a}}, J., and {Haji{\v{c}}}, J.
\newblock (2019b).
\newblock {Evaluating Contextualized Embeddings on 54 Languages in POS Tagging,
  Lemmatization and Dependency Parsing}.
\newblock {\em arXiv e-prints}, page arXiv:1908.07448, August.

\bibitem[\protect\citename{Straka}2018]{UDPipe2018}
Straka, M.
\newblock (2018).
\newblock {UDPipe 2.0 Prototype at CoNLL 2018 UD Shared Task}.
\newblock In {\em Proceedings of CoNLL 2018: The SIGNLL Conference on
  Computational Natural Language Learning}, pages 197--207, Stroudsburg, PA,
  USA. Association for Computational Linguistics.

\bibitem[\protect\citename{Stymne \bgroup et al.\egroup
  }2018]{stymne-etal-2018-parser}
Stymne, S., de~Lhoneux, M., Smith, A., and Nivre, J.
\newblock (2018).
\newblock Parser training with heterogeneous treebanks.
\newblock In {\em Proceedings of the 56th Annual Meeting of the Association for
  Computational Linguistics (Volume 2: Short Papers)}, pages 619--625,
  Melbourne, Australia, July. Association for Computational Linguistics.

\bibitem[\protect\citename{Vaswani \bgroup et al.\egroup }2017]{vaswani:2017}
Vaswani, A., Shazeer, N., Parmar, N., Uszkoreit, J., Jones, L., Gomez, A.~N.,
  Kaiser, L., and Polosukhin, I.
\newblock (2017).
\newblock Attention is all you need.
\newblock {\em CoRR}, abs/1706.03762.

\bibitem[\protect\citename{Zeman \bgroup et al.\egroup
  }2017]{UDST2017:overview}
Zeman, D., Popel, M., Straka, M., Haji{\v{c}}, J., Nivre, J., Ginter, F.,
  et~al.
\newblock (2017).
\newblock {CoNLL 2017 Shared Task: Multilingual Parsing from Raw Text to
  Universal Dependencies}.
\newblock In {\em {Proceedings of the CoNLL 2017 Shared Task: Multilingual
  Parsing from Raw Text to Universal Dependencies}}, pages 1--19, Vancouver,
  Canada. Association for Computational Linguistics.

\bibitem[\protect\citename{Zeman \bgroup et al.\egroup
  }2018]{UDST2018:overview}
Zeman, D., Haji{\v{c}}, J., Popel, M., Potthast, M., Straka, M., Ginter, F.,
  Nivre, J., and Petrov, S.
\newblock (2018).
\newblock {CoNLL 2018 Shared Task: Multilingual Parsing from Raw Text to
  Universal Dependencies}.
\newblock In {\em Proceedings of the CoNLL 2018 Shared Task: Multilingual
  Parsing from Raw Text to Universal Dependencies}, pages 1--20, Brussels,
  Belgium, October. Association for Computational Linguistics.

\end{thebibliography}


\begin{thebibliography}{}

\bibitem[\protect\citename{Ginter \bgroup et al.\egroup
  }2017]{CoNLL2017RawTexts}
Ginter, F., Haji{\v c}, J., Luotolahti, J., Straka, M., and Zeman, D.
\newblock (2017).
\newblock {\em {CoNLL} 2017 Shared Task - Automatically Annotated Raw Texts and
  Word Embeddings}.
\newblock Institute of Formal and Applied Linguistics, LINDAT/CLARIN, Charles
  University, Prague, Czech Republic, LINDAT/CLARIN PID:
  http://hdl.handle.net/11234/1-1989.

\bibitem[\protect\citename{Zeman \bgroup et al.\egroup }2019]{UD2.5}
Zeman, D., Nivre, J., et~al.
\newblock (2019).
\newblock {\em Universal Dependencies 2.5}.
\newblock Institute of Formal and Applied Linguistics, LINDAT/CLARIN, Charles
  University, Prague, Czech Republic, LINDAT/CLARIN PID:
  http://hdl.handle.net/11234/1-3105.

\end{thebibliography}

\section{Language Resource References}
\bibliographystylelanguageresource{lrec}
\bibliographylanguageresource{evalatin_udpipe}

\end{document}